\RequirePackage{fix-cm}
\documentclass[twocolumn,compsoc]{cvm}

\graphicspath{{figures/},{figures/cvm/}}

\def\ManuscriptInfo{Manuscript received: 2021-08-08; accepted: xxxx-xx-xx.}

\def\PaperTitle{Dynamic Hypergraph Convolutional Networks for Skeleton-Based Action Recognition}

\def\AuthorNames{Jinfeng Wei$^1$, Yunxin Wang$^1$, Mengli Guo$^1$, Pei Lv$^1$(\rm{\Letter}), \textbf{Xiaoshan Yang$^2$, and Mingliang Xu$^1$}}

\def\AuthorAdress{$1\quad $ & Zhengzhou University, Henan 450001, China. E-mail: 

            J. Wei, wjf1997@gs.zzu.edu.cn; 
            
            Y. Wang, wangyunxin@gs.zzu.edu.cn; 
    
            M. Guo, molly957100975@gs.zzu.edu.cn; 
            
            P. Lv, ielvpei@zzu.edu.cn (\rm{\Letter});
            
            M. Xu, iexumingliang@zzu.edu.cn.\\
            $2\quad $ & National Lab of Pattern Recognition, Institute of Automation, Chinese Academy of Sciences, Beijing 100190, China, E-mail: xiaoshan.yang@nlpr.ia.ac.cn.\\
}

\def\firstheader{DOI 10.1007/s41095-xxx-xxxx-x\hfill Vol. x, No. x, month year, xx--xx\\[5mm]~Article Type (Research)}

\headevenname{{Jinfeng Wei \etal} }

\begin{document}

\MakePageStyle

\MakeAbstract{Graph convolutional networks (GCNs) based methods have achieved advanced performance on skeleton-based action recognition task. However, the skeleton graph cannot fully represent the motion information contained in skeleton data. In addition, the topology of the skeleton graph in the GCN-based methods is manually set according to natural connections, and it is fixed for all samples, which cannot well adapt to different situations. In this work, we propose a novel dynamic hypergraph convolutional networks (DHGCN) for skeleton-based action recognition. DHGCN uses hypergraph to represent the skeleton structure to effectively exploit the motion information contained in human joints. Each joint in the skeleton hypergraph is dynamically assigned the corresponding weight according to its moving, and the hypergraph topology in our model can be dynamically adjusted to different samples according to the relationship between the joints. Experimental results demonstrate that the performance of our model achieves competitive performance on three datasets: Kinetics-Skeleton 400, NTU RGB+D 60, and NTU RGB+D 120.
}

\MakeKeywords{spatial temporal convolutional network; \ skeleton-based action recognition; \ dynamic hypergraph; \ skeleton data}

\section{Introduction}\label{sec:introduction}
\noindent Action recognition has important applications in virtual reality \cite{ref_30} and intelligent surveillance \cite{ref_31}. With the popularization of 3D depth cameras and the development of human pose estimation technology \cite{ref_23}, human skeleton data has become easier to obtain. Compared with traditional action recognition methods using RGB video sequences, skeleton data can retain high-order motion information. Skeleton-based action recognition methods are more robust to influencing factors such as illumination, color and occlusion.

The previous methods \cite{ref_1,ref_2,ref_3} mainly use hand-crafted features to complete the action recognition task. Subsequently, action recognition methods based on recurrent neural networks (RNNs) \cite{ref_10,ref_11,ref_12,ref_13,ref_14,ref_16} and convolutional neural networks (CNNs) \cite{ref_4,ref_5,ref_6,ref_7,ref_8,ref_9} have gradually become the mainstream. They usually take the  human joint coordinates as a vector sequence or a pseudo-image into RNNs or CNNs to obtain action recognition results. However, vector sequences or pseudo-images cannot well demonstrate the graph structure of skeleton data. Therefore, these methods cannot fully extract the features contained in the skeleton data.

Recently, graph convolutional networks (GCNs) have been used in action recognition \cite{ref_17,ref_18}. Yan et al. \cite{ref_17} firstly used GCN to model the skeleton data and proposed a spatial-temporal graph convolutional network (ST-GCN). Shi et al. \cite{ref_18} proposed to add a learnable matrix into the normalized matrix of the skeleton graph to improve the adaptability of the model. These methods are built on the basis of graph convolutional networks. However, the skeleton graph cannot fully represent the human skeleton structure, and this inherent shortcoming cannot be solved by simply adjusting the topology of the graph. In the 3D pose estimation task, Liu et al. \cite{ref_32} proposed a semi-dynamic hypergraph neural network, which used one static hypergraph to represent the tree body structure of human, and this method can obtain more information than the traditional graph structure.

\begin{figure*}[!ht]
\centering
\subfigure[]{\includegraphics[width=0.18\textwidth]{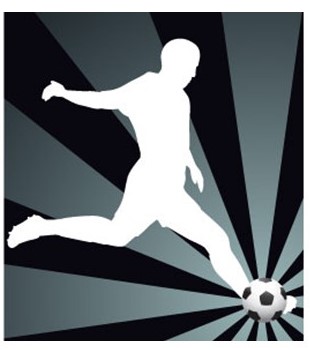}\label{fig1:a}} 
\subfigure[]{\includegraphics[width=0.18\textwidth]{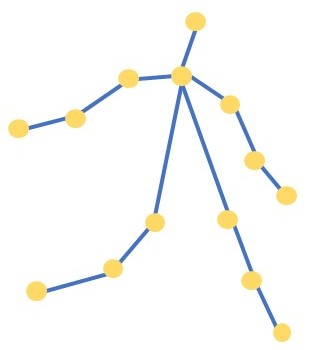}\label{fig1:b}} 
\subfigure[]{\includegraphics[width=0.18\textwidth]{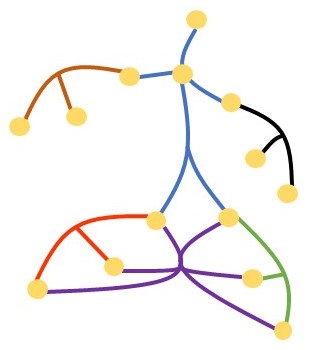}\label{fig1:c} } 
\subfigure[]{\includegraphics[width=0.18\textwidth]{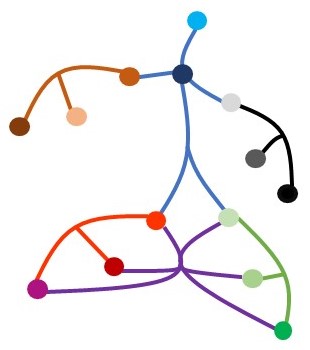}\label{fig1:d}} 
\subfigure[]{\includegraphics[width=0.18\textwidth]{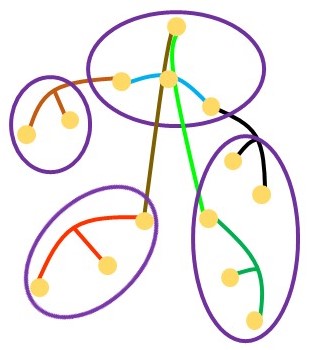}\label{fig1:e}} 
\caption{Human skeleton representation based on graph and hypergraph. (a) shows someone is kicking soccer ball, which is a simple of skeleton data. (b) shows the traditional skeleton graph. (c) shows our static hypergraph with six hyperedges. The lines with different colors represent hyperedges, and the nodes on the lines represent the human body joints. (d) shows our dynamic joint weight of hypergraph, each joint is assigned the corresponding weight with different colors. (e) shows our dynamic topology of hypergraph. The lines with different colors represent the hyperedges obtained by the $K$-NN method, and the ellipses represent the hyperedges containing the global information obtained by the $K$-means method.}
\label{fig1}
\end{figure*}

Although GCN-based method is proven effective in skeleton-based action recognition, it still has three obvious shortcomings: (1) The edges in the graph structure can only connect two related nodes, but the joints of the human skeleton are not simply pairwise connections. Therefore, the graph structure cannot well capture the high-order information hidden between multiple joints. (2) Previous GCN-based methods mainly use skeleton data by modeling the natural connections of the human skeleton, however, they ignore indirect connections such as hands and legs (\figref{fig1:b}). 
(3)Each dataset contains dozens or even hundreds of action categories, and each action category contain thousands of samples. The position and angle of human joints in these samples will be very different, but the traditional skeleton graph structure is fixed for all samples, which cannot be flexibly adjusted to meet the needs of various situations.

In this paper, we propose a novel dynamic hypergraph convolutional networks (DHGCN) to solve the above problems. In our method, hypergraph is used to model the skeleton data and we design a dynamic hypergraph, which is composed of two parts: dynamic joint weight and dynamic topology. For the dynamic joint weight of each joint, we use the moving distance of one to estimate it (\figref{fig1:d}). This method can assign corresponding weights to joints in the skeleton hypergraph, so that the skeleton hypergraph can reflect the importance of each joint to help action recognition. For the dynamic topology, we use the $K$-NN method and $K$-means method to construct the topology of different samples (\figref{fig1:e}). This method can better extract the motion features of human joints, and generate more suitable topology for different samples. The proposed dynamic hypergraph can improve the flexibility of the model. In addition, one static hypergraph is used to represent the basic topology of the original skeleton structure (\figref{fig1:c}). These two hypergraphs are combined together to extract the spatial features contained in the skeleton data. Afterwords, the spatial information is processed by the temporal convolutional network (TCN) to further obtain the spatial-temporal information of the skeleton data. In order to validate the effectiveness of our proposed method, we conduct experiments on three datasets: Kinetics-Skeleton 400 \cite{ref_21}, NTU RGB+D 60 \cite{ref_11}, and NTU RGB+D 120 \cite{ref_43}.

The contributions of our work is mainly in three aspects:
\begin{itemize}
\item[-]A dynamic hypergraph convolutional networks (DHGCN) for skeleton-based action recognition is proposed. The skeleton data is modeled by hypergraph, and we use a combination of static hypergraph and dynamic hypergraph to capture the motion information.

\item[-]The proposed dynamic hypergraph is composed of two parts: dynamic joint weight and dynamic topology. Dynamic joint weight is proposed to assign each joint corresponding weight, which can better help to aggregate the node features. The dynamic topology of the skeleton hypergraph of different samples learned in an end-to-end manner.

\item[-]Our proposed DHGCN achieves competitive performance on Kinetics-Skeleton 400, NTU RGB+D 60, and NTU RGB+D 120, and the experiments prove its superiority.
\end{itemize}

\section{Related work} \label{sec:Relatedwork}

\subsection{Skeleton-based action recognition}
\noindent Traditional skeleton-based action recognition methods \cite{ref_1,ref_2,ref_3} mainly tend to use hand-crafted features to model the skeleton data and classify actions according to joint trajectories. They use traditional machine learning methods to extract features from the video and encode the features, then normalize the encoding vector, and finally train the model to obtain the prediction result. For example, Vemulapalli et al. \cite{ref_2} used the relative geometric position of joints to model skeleton data. The advantage of this kind of method is that it can extract features according to different needs and it is easy to implement. However, the hand-crafted features are not well suitable to the action classification task since it does not fully consider the influencing factors at a certain moment. 

Subsequently, the methods based on RNNs and CNNs are beginning to be used for action recognition. RNN-based methods \cite{ref_10,ref_11,ref_12,ref_13,ref_14,ref_15,ref_16} usually use coordinate vectors to model the skeleton data, then the obtained sequence is input to the RNN-based model to obtain the prediction result. CNN-based methods \cite{ref_4,ref_5,ref_6,ref_7,ref_8,ref_9} use a pseudo-image to model the skeleton data. For example, Ke et al. \cite{ref_9} used deep CNN to process the input sequence to extract CNN features. The CNN-based methods are easier to train because they have better parallelism. However, these methods based on RNNs and CNNs cannot fully express the information contained in skeleton data because vector sequences or pseudo-images cannot well demonstrate the graph structure of skeleton data. 

\subsection{GCN-based action recognition}
\noindent Recently, the usage of GCN in modeling the human skeleton data has provided a new direction for solving action recognition task. Yan et al. \cite{ref_17} proposed a spatial-temporal graph convolutional network (ST-GCN), which is the first GCN-based method to model the skeleton data and this method successfully extracted the features contained in the skeleton data. Shi et al. \cite{ref_18} proposed an adaptive graph, which made the skeleton graph more flexible. This method added a parameter matrix completely learned from the model to the feature matrix of GCN. This matrix represented the possible joint relationship in the movement process and they proposed a two-stream framework to improve the recognition accuracy. Zhang et al. \cite{ref_19} proposed context aware graph convolution, which integrated long range dependencies between joints into context information. This method reduced the number of parameters and the depth of the network. Shi et al. \cite{ref_22} firstly used directed acyclic graph to model the skeleton data. However, the skeleton graph in these methods cannot meet the needs of different actions, because the relationship between the joints in different samples is not fixed.

\subsection{Hypergraph neural networks}
\noindent Hypergraph neural networks \cite{ref_20} is mainly used for data representation learning. The main problem it was designed to solve is: the relationship of data may be more complicated than the pairwise connections, and is difficult to model with common graph structure. Hypergraph neural network can make better use of higher-order data correlation for representation learning. Jiang et al. \cite{ref_29} proposed dynamic hypergraph neural networks, each layer of the network dynamically updates the hypergraph topology. Yadati et al. \cite{ref_34} proposed hypergraph convolutional networks, which is a new method of training hypergraph networks. Gao et al. \cite{ref_37} proposed a dynamic hypergraph representation method and a new learning framework. In skeleton-based action recognition task, there are multiple joints that need to be coordinated when completing an action. In this case, the connections of the skeleton data is represented by a hypergraph structure, which can better extract the features between multiple joints.

\section{Methods}\label{sec:method}

\noindent In this section, we first briefly introduce the traditional GCN and the concept of hypergraph convolutional network (HGCN). Then we show the details of the proposed dynamic hypergraph and the complete network structure based on it.

\begin{figure}[!ht]
\includegraphics[width=0.5\textwidth]{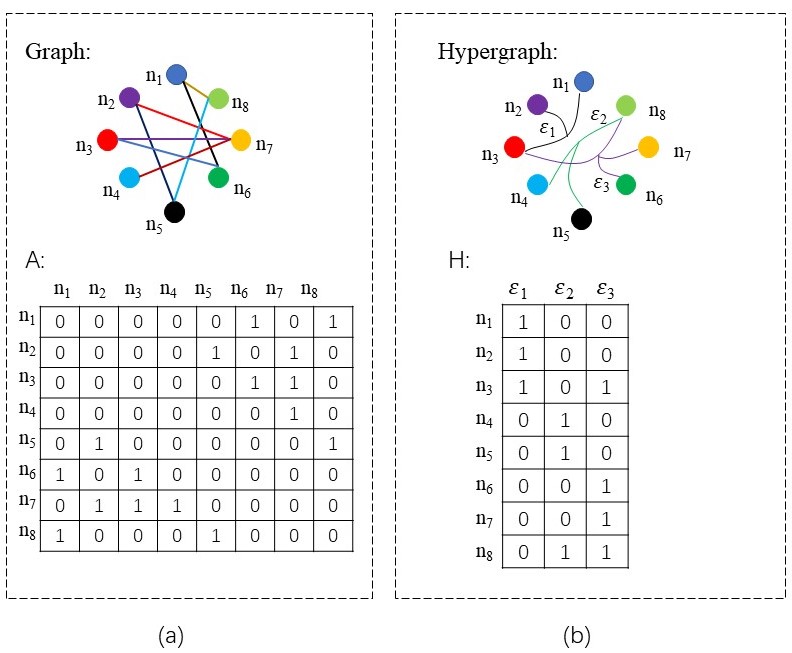}
\caption{The feature matrix of graph convolutional networks and hypergraph convolutional networks.} \label{fig2}
\end{figure}

\subsection{GCN}
\noindent The traditional GCN-based methods use $G=\{V,E\}$ to represent the human skeleton graph, where $V$ represents the joint set, $E$ represents the edge set, which consists of $E_S$ and $E_T$. $E_S=\{(v_{ti},v_{tj})|i,j=1,...,V,t=1,...,T\}$ means that at time step $t$, each pair of joints $(v_{ti},v_{tj})$ corresponding to the bones is connected. Another subset $E_T=\{(v_{ti},v_{(t+1)j})|i=1,...,V,t=1,...,T\}$ represents the connection of the same joint among all frames. The set of joints and the set of edges together form a skeleton graph to model the skeleton data.

The feature matrix of the skeleton graph is represented by $A$. If two joints are on an edge, the value of the element in $A$ corresponding to the two joints is equal to $1$, otherwise it is equal to $0$ (\figref{fig2}(a)). The update rule of graph convolutional networks at time step $t$ can be defined as:
\begin{equation}
    X_t^{(l+1)}=\sigma(\tilde{D}^{-\frac{1}{2}}\tilde{A}\tilde{D}^{-\frac{1}{2}}X_t^{(l)}\Theta^{(l)})
\end{equation}
where $\sigma(\cdot)$ is a non-liner activation function, $\tilde{D}$ is the degree matrix of the node, $X_t^{(l)}$ is the input data of the convolutional layer $l$ at time step $t$, $\Theta^{(l)}$ is the learnable matrix of the convolutional layer $l$, $\tilde{A}$ is a normalized graph adjacency matrix. Here, $\tilde{A}=A+I$, and $I$ is an identity matrix.

\begin{figure*}[!ht]
\includegraphics[width=\textwidth]{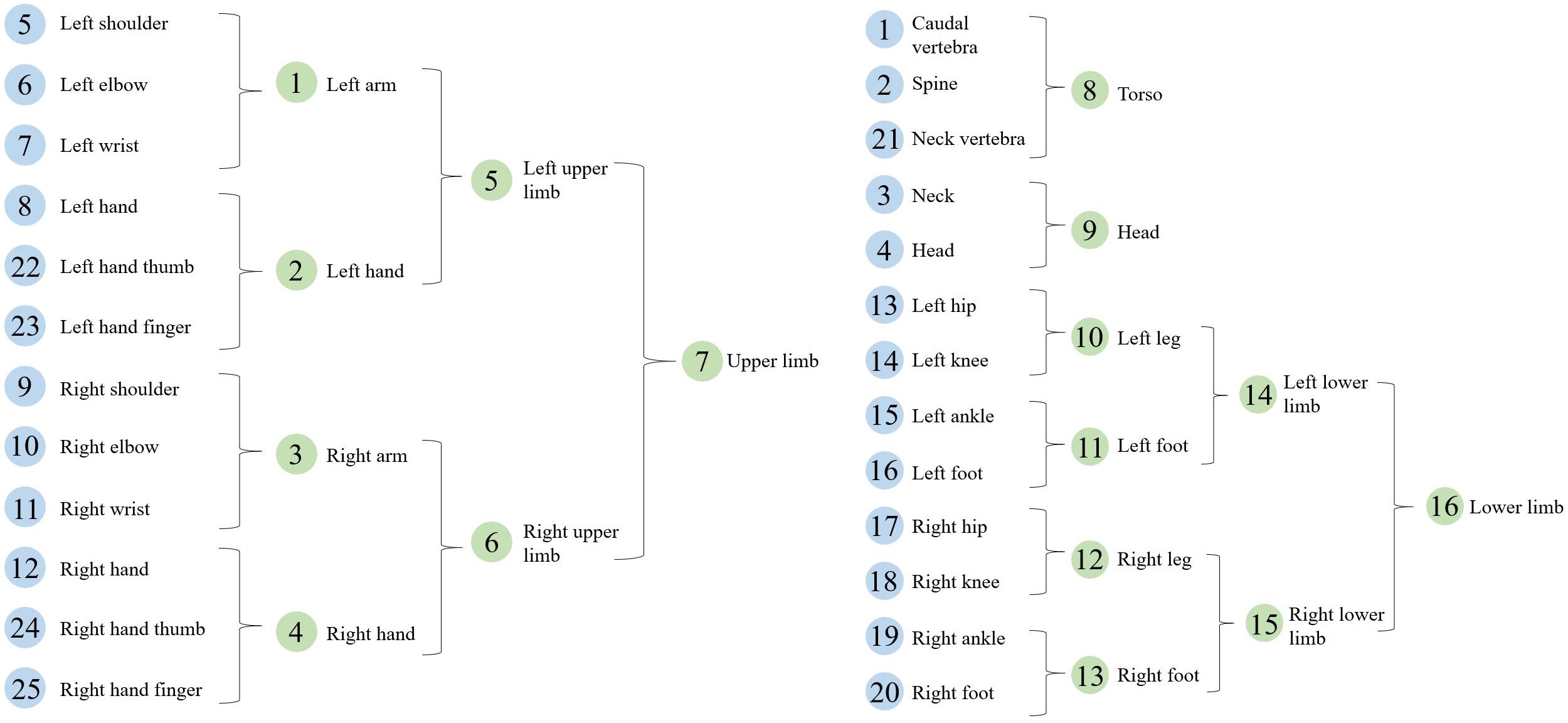}
\caption{The static hypergraph for NTU RGB+D 60 dataset. The blue dots represent joints, and the green dots represent hyperedges composed of multiple joints.} \label{fig3}
\end{figure*}

\subsection{HGCN} 
\noindent We update the graph in above method with hypergraph, and the human skeleton hypergraph is represented by $G_h=\{V_h,\xi_h,W_h\}$, where $V_h$ represents the set of all joints in one frame, $\xi_h$ is the hyperedge set, $W_h$ is the set of weights of all hyperedges. In our work, we use a hyperedge to connect multiple related joints, including unnatural connected joints such as hands and legs. The hyperedges of our static hypergraph are shown in \figref{fig3}.

The elements in the incident matrix $H$ of this skeleton hypergraph (\figref{fig2}(b)) are defined as:
\begin{equation}
    H=\left\{
\begin{aligned}
h(v,e)=1 \qquad v\in{V_h},e\in \xi_h,v\in e \\
h(v,e)=0 \qquad v\in{V_h},e\in \xi_h,v\notin e
\end{aligned}
\right.
\end{equation}

\noindent The degree of node $v\in{V_h}$ is the number of hyperedges containing this node, which follows:
\begin{equation}
    d(v)=\sum_{e\in \xi_h}W_h(e)h(v,e)
\end{equation}
where $W_h(e)$ represents the weight of hyperedge $e$. The degree of hyperedge $e\in \xi_h$ is the number of joints constituting the hyperedge $e$, which follows:
\begin{equation}
    \delta(e)=\sum_{v\in{V_h}}h(v,e)
\end{equation}
The update rule of the hypergraph convolutional networks at time step $t$ can be generalized by the convolution formula in \cite{ref_20}:
\begin{equation}\label{eq2}
    X_t^{(l+1)}=\sigma(D_v^{1/2}HWD_e^{-1}H^TD_v^{1/2}X_t^{(l)}\Theta^{(l)})
\end{equation}
where $D_v$, $D_e$ denote the diagonal matrices of the degrees of nodes and the degrees of the hyperedges, $W$ is the diagonal matrix of the weights of all hyperedges. Initially, the weights of all the hyperedges are set to 1. $\Theta^{(l)}$ is a parameter matrix learned during training process.

\begin{figure*}[!ht]
\includegraphics[width=\textwidth]{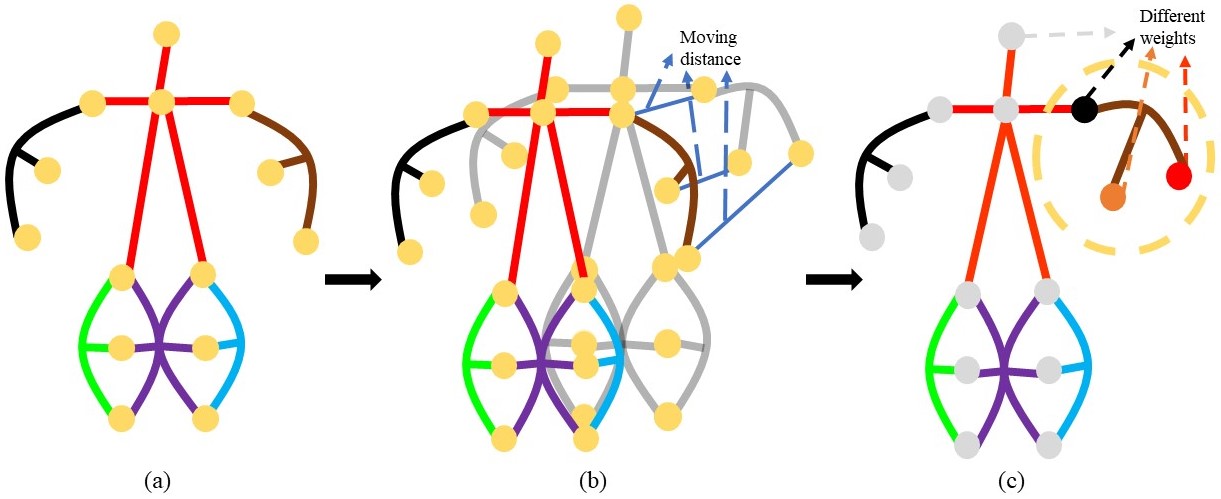}
\caption{The illustration of deriving the weight of each joint when the hand is raised in the part of dynamic joint weight. (a) The spatial position of each joint at time step $t$. (b) The blue line represents the distance that the corresponding joint moves in two frames. (c) Each joint is assigned a weight with different colors according to the moving distance.} \label{fig4}
\end{figure*}

\subsection{Dynamic joint weight of hypergraph}
\noindent The process of hypergraph convolution has two steps. First, the features of all the nodes belonging to the hyperedge $e$ are aggregated to obtain the feature of the hyperedge $e$, then all the hyperedge features containing node $v$ are aggregated to obtain the feature of node $v$. We calculate different weights of various joints during above two aggregation processes at each time step, which can extract richer action features. 

The weight of each joint is obtained according to its moving distance (\figref{fig4}). The moving distance of one joint is obtained by firstly calculating the difference between the corresponding joint coordinates in two adjacent frames, and then calculating the second norm of the displacement vector to obtain the displacement $dis_{v_i^t}$ of the same joint between the two frames:
\begin{equation}
    dis_{v_i^{t}}=||v_i^{t}-v_i^{t-1}||_2
\end{equation}
where $v_i^{t}$ represents the three-dimensional coordinates of joint $v_i$ at time step $t$, $v_i^{t-1}$ represents the three-dimensional coordinates of joint $v_i$ at time step $t-1$.
At time step $t$, the weight $W_{v_i^t}$ of node $v_i^{t}$ on each hyperedge $e\in{\xi_h}$ can be calculated:
\begin{equation}
    W_{v_i^{t}}=softmax(dis(v_i^{t}))=\frac{dis_{v_i^{t}}}{\sum\limits_{{v_j^{t}\in{e_k^{t}}}}^{}dis_{v_j^{t}}}
\end{equation}
where $e_k^{t}$ represents the hyperedge containing node $v_i$, $v_j^{t}$ represents the nodes on hyperedge $e_k^{t}$. After obtaining the weight matrix $W_{all}$ of all nodes, we combine $W_{all}$ with the incident matrix $H$ to obtain the hypergraph convolution operator $Imp$, which follows:

\begin{equation}
    Imp=W_{all} \cdot H
\end{equation}
then we change \eqref{eq2} to the following formula:
\begin{equation}
    X_t^{(l+1)}=\sigma(ImpImp^TX_t^{(l)}\Theta^{(l)})
\end{equation}

\subsection{Dynamic topology of hypergraph}\label{sec:topology}
\begin{figure*}[!ht]
\centering
\includegraphics[width=\textwidth]{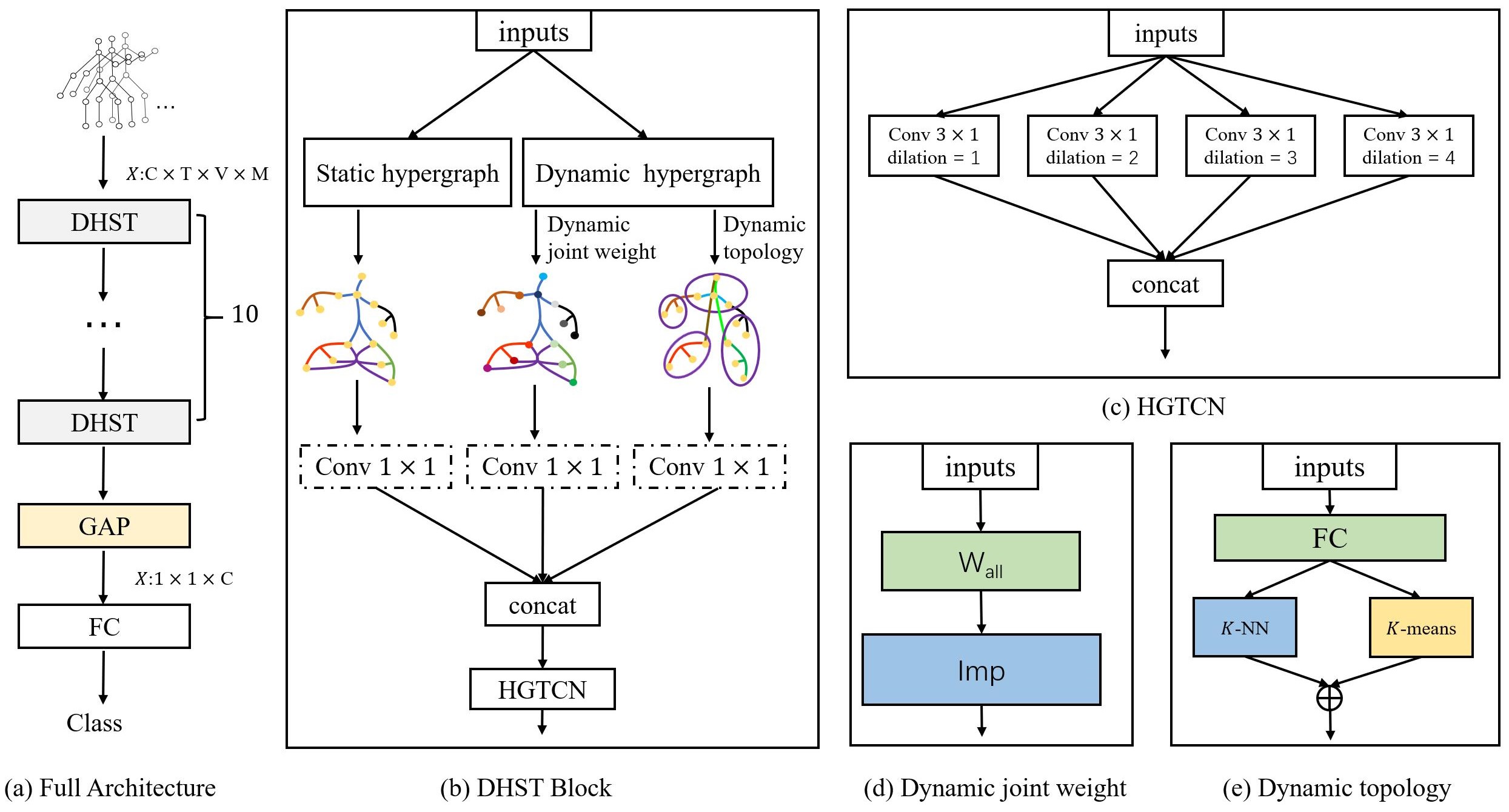}
\caption{Architecture overview. The whole model includes $10$ DHST blocks.} \label{fig5}
\end{figure*}

\noindent We dynamically change the topology of the skeleton hypergraph to improve the flexibility of the model (\figref{fig5}(e)). After the skeleton data is processed by the fully connected (FC) layer, we can obtain a new node feature vector set $X_{new}$, which follows:
\begin{equation}
    X_{new} = \sigma(W_{map}f_{in})
\end{equation}
where $f_{in}$ represents the input features, $W_{map}$ represents the mapping between input features and output features, which is a learnable matrix.

After the skeleton data is mapped by the FC layer, the distance between related nodes will be close \cite{ref_36}. We use the $K$-NN method and $K$-means method to obtain two sets of hyperedges: common information hyperedge set and global information hyperedge set. The hyperedges in the union of these two sets constitute a new skeleton hypergraph.

Each frame of skeleton data contains $N$ joints. For these $N$ joints, we calculate the Euclidean distance between joint $v_i$ and all other joints at time step t, which follows:
\begin{equation}
\begin{aligned}
    D_{v_i}^t\!=\! \sqrt{{{({X_{v_i}^t}\!-\!{X_{v_j}^t})}^2}\!+\!{{({Y_{v_i}^t}\!-\!{Y_{v_j}^t})}^2}\!+\!{{({Z_{v_i}^t}\!-\!{Z_{v_j}^t})}^2}}
\end{aligned}
\end{equation}
where $v_i$ and $v_j$ represent different joints.

For each joint, we connect the $k_n$ joints with the smallest distance to form a hyperedge $e$. $k_n$ determines the number of joints are contained in each hyperedge. Finally, a set containing $N$ hyperedges and $k_n$ nodes on each hyperedge is obtained. 

Afterwards, we randomly select $k_m$ joints from $N$ joints in one frame as the centroids, all nodes are divided into $k_m$ sets according to the distance from other nodes to the centroids. Then, the node with the smallest mean value of the distance to other nodes in one set is taken as the new centroid, and the position of the centroid is iteratively updated according to this method until the change of the position of the centroid is $0$. $k_m$ determines the number of hyperedges. Finally, we divide the joints of each frame into $k_m$ disjoint sets, and the joints in each set are connected to form a hyperedge, and then a set containing $k_m$ hyperedges is obtained. These hyperedges fully contain the global information of the skeleton data.

\subsection{DHGCN}\label{sec:model}
\noindent The whole model uses ten layers of backbone called Dynamic Hypergraph Spatial-Temporal Convolution Block (DHST), which includes spatial convolution and temporal convolution (\figref{fig5}). The spatial convolution is used to obtain the spatial information between joints and bones, and the temporal convolution is used to capture the temporal information embedded in different frames. We integrate the features of three branches in the spatial convolution module, including static hypergraph, dynamic joint weight and dynamic topology. The convolution kernel size of the temporal convolution module is fixed at $3\times1$, and a larger receptive field can be obtained by using different dilation rates. The extracted spatial-temporal features are processed by the global average pooling (GAP) layer and the FC layer to obtain prediction scores. Human action information is not simply contained in the joints, but the bones also contain action information. Both the lengths and the angles of the bones contain rich information which can help to recognize the action. Therefore, we use a two-stream framework of joint-bone fusion inspired by 2s-AGCN \cite{ref_18}. The proposed model is used to train joint data and bone data respectively, and the prediction scores of the bone model and joint model are summed to obtain the final classification result.

\section{Results and discussion}\label{sec:Results}

\subsection{Datasets}
\noindent Kinetics Skeleton 400 \cite{ref_21} dataset contains about 300,000 video clips collected from YouTube, which includes up to 400 human actions. These actions involve human daily activities, sports scenes and complex human-computer interaction scenes. However, the Kinetics dataset only provides raw videos without skeleton sequences. ST-GCN uses the public OpenPose toolbox \cite{ref_23} to obtain the positions of human body joints on each frame. We use the skeleton data (Kinetics-Skeleton) obtained after their processing to evaluate our model. To be consistent with the method in \cite{ref_17}, we train the model and report the Top-1 and Top-5 accuracies on the test set. 

NTU RGB+D 60 \cite{ref_11} is a large dataset containing 56578 skeleton sequences. It was collected by 40 subjects and shot by three cameras with the same height but different horizontal angles. It has 60 action categories. Each skeleton sequence has 25 joints, and there are no more than two people in each video. There are two evaluation criteria for this dataset: (1) Cross-subject (X-Sub): 40 subjects are divided into training and testing set. (2) Cross-view (X-View): all skeleton sequences collected from camera 1 as test set, and the other skeleton sequences as train set.

NTU RGB+D 120 \cite{ref_43} is the largest skeleton action recognition dataset, and 60 action categories are added on the basis of NTU RGB+D 60. There are two evaluation criteria for this dataset: (1) Cross-subject (X-Sub): 106 subjects are divided into training and testing set. (2) Cross-Setup (X-Set): samples with even IDs are used as training samples, and
samples with odd IDs are used as testing data.

\subsection{Training details}
\noindent All experiments are performed on the PyTorch deep learning framework and we use SGD with momentum $0.9$ as the optimizer. The batch size is 16. We use cross-entropy as the loss function. Initially, the learning rate is set to $0.1$. Learning rate is divided by 10 when the learning rate decays. When training on the NTU RGB+D 60 and NTU RGB+D 120, the learning rate decays at the $30_{th}$ and $40_{th}$ epoch and the training process ends at the $50_{th}$ epoch. For Kinetics-Skeleton, the learning rate decays at the $45_{th}$ and $55_{th}$ epoch and the training process ends at the $65_{th}$ epoch.
\subsection{Ablation study}
\noindent In order to prove that skeleton hypergraph performs better than skeleton graph in skeleton-based action recognition task, we select 2s-AGCN as the baseline and replace the graph convolutional networks with the hypergraph convolutional networks to obtain 2s-AHGCN. 2s-AHGCN is evaluated on Kinetics-Skeleton and NTU RGB+D 60 datasets, and the comparison of the results is shown in \tabref{tab1}.

\begin{table}[!h]
\caption{The effectiveness of hypergraph on existing GCN-based method.}\label{tab1}
\begin{tabular*}{\hsize}{@{}@{\extracolsep{\fill}}lcccc@{}}
\toprule
\multirow{2}{*}{Methods} & \multicolumn{2}{c}{Kinetics-Skeleton}& \multicolumn{2}{c}{NTU RGB+D 60}  \\ \cline{2-5}
&Top1&Top5&X-Sub&X-View\\
\hline
2s-AGCN(Joint) &  35.1 & 57.1 &- & 93.7\\
2s-AHGCN(Joint) & 35.5 & 57.6 & 87.5 & 94.2\\
\hline
2s-AGCN(Bone) &  33.3 & 55.7 &- & 93.2\\
2s-AHGCN(Bone) & 34.5 & 56.8 & 87.6 & 93.6\\
\hline
2s-AGCN &  36.1 & 58.7 &88.5 & 95.1\\
2s-AHGCN & \textbf{37.0} & \textbf{59.8} & \textbf{89.4} & \textbf{95.4}\\
\bottomrule
\end{tabular*}
\end{table}

The Top-1 accuracy of 2s-AHGCN on the Kinetics-Skeleton dataset has increased by 0.9\%, and the Top-5 accuracy has increased by 1.1\%. The X-Sub accuracy of 2s-AHGCN on the NTU RGB+D 60 dataset has increased by 0.9\%, and the X-View accuracy has increased by 0.3\%. The performance of 2s-AHGCN on the two datasets demonstrate that the hypergraph can better model the skeleton data.

Thakkar et al. \cite{ref_42} proposed PB-GCN, which divided the original human skeleton graph into multiple subgraphs. Their model performs convolution operations on each subgraph, and an aggregation function is used to aggregate the features of all subgraphs. We use the subgraphs in PB-GCN as hyperedges to construct a new model PB-HGCN, which eliminates the need of aggregation functions. Experiment results on the NTU RGB+D 60 dataset prove that our hypergraph convolution is better than PB-GCN.

\begin{table}[!h]
\caption{ Ablation study of different numbers of subgraphs on NTU RGB+D 60.}\label{tab3}
\begin{tabular*}{\hsize}{@{}@{\extracolsep{\fill}}lcc@{}}
\toprule
\multirow{2}{*}{Methods} & \multicolumn{2}{c}{NTU RGB+D 60}  \\ \cline{2-3}
&X-Sub&X-View\\
\hline
PB-GCN(two) & 80.2 &88.4 \\
PB-HGCN(two) & 81.6 &90.2 \\
PB-GCN(four) & 82.8 &90.3 \\
PB-HGCN(four) & \textbf{84.9} & \textbf{91.7} \\
PB-GCN(six) & 81.4 &89.1 \\
PB-HGCN(six) & 82.5 &90.8 \\
\bottomrule
\end{tabular*}
\end{table}

In \secref{sec:topology}, we use the $K$-NN method and $K$-means method to obtain the dynamic topology of hypergraph. $k_n$ determines the number of joints contained in each hyperedge of the common information hyperedge set. $k_m$ determines the number of hyperedges containing the global information. In order to select the appropriate $k_n$ and $k_m$, we set different values for them to conduct comparative experiments to find the best configuration.

\begin{table}[!h]\setlength\tabcolsep{2pt}%
\caption{ The performance of our model with different settings.}\label{tab2}
\begin{tabular*}{\hsize}{@{}@{\extracolsep{\fill}}lcccc@{}}
\toprule
\multirow{2}{*}{Methods} & \multicolumn{2}{c}{Kinetics-Skeleton}& \multicolumn{2}{c}{NTU RGB+D 60}  \\ \cline{2-5}
&Top1&Top5&X-Sub&X-View\\
\hline
DHGCN($k_n$=2,$k_m$=3) &  37.0 & 59.6 &90.1 & 95.1\\
DHGCN($k_n$=2,$k_m$=4) &  37.2 & 60.1 &90.3 & 95.4\\
DHGCN($k_n$=2,$k_m$=5) & 36.8 & 59.7 & 90.1 & 95.2\\
DHGCN($k_n$=3,$k_m$=3) &  37.2 & 60.2 &90.3 & 95.6\\
DHGCN($k_n$=4,$k_m$=3) &  36.9 & 59.7 &90.0 & 95.2\\
DHGCN($k_n$=3,$k_m$=4) &  \textbf{37.7} & \textbf{60.6} &\textbf{90.7} & \textbf{96.0}\\
\bottomrule
\end{tabular*}
\end{table}

The results in \tabref{tab2} show that when $k_n$ is greater than 3 or $k_m$ is greater than 4, the performance of DHGCN begins to decline. The model achieves the best performance when $k_n$ is set to 3 and $k_m$ is set to 4. The influence of $k_n$ on the performance of DHGCN proves that there is a threshold for the number of joints constituting a hyperedge. When $k_n$ exceeds the threshold, the performance of DHGCN will gradually decrease. The influence of $k_m$ on the performance of DHGCN proves that when the number of hyperedges containing global information reaches to a threshold, if $k_m$ continues to increase, the number of joints contained in each hyperedge will be too small, which leads to these hyperedges cannot well reflect the global information among joints. The DHGCN mentioned later are all set to $k_n$ equal to 3 and $k_m$ equal to 4.

In \secref{sec:model}, we propose a dynamic hypergraph convolutional network. There are three different branches in the spatial convolution module, and we try to delete one of them and evaluate their performance on NTU RGB+D 60 dataset. \tabref{tab3} shows that dynamic hypergraph convolutional network is beneficial for action recognition task. Removing any one of these three branches will decrease the performance of the model. Especially after deleting dynamic joint weight and dynamic topology, the performance of the model is greatly reduced, which proves that our proposed dynamic hypergraph plays an important role. When these three branches are combined, the performance of DHGCN is the best.

\begin{table}[!h]
\caption{ The performance comparison of dynamic hypergraph convolutional networks with or without static hypergraph, dynamic joint weight and dynamic topology. no/$X$ means not included the $X$ module.}\label{tab4}
\begin{tabular*}{\hsize}{@{}@{\extracolsep{\fill}}lcc@{}}
\toprule
\multirow{2}{*}{Methods} & \multicolumn{2}{c}{NTU RGB+D 60}  \\ \cline{2-3}
&X-Sub&X-View\\
\hline
DHGCN(no/static) & 90.3 &95.6 \\
DHGCN(no/joint) & 90.0 &95.1 \\
DHGCN(no/topology) & 89.9 &94.7 \\
DHGCN(no/dynamic) & 88.7 &94.3 \\
DHGCN & \textbf{90.7} & \textbf{96.0}\\
\bottomrule
\end{tabular*}
\end{table}

We also verify our two-stream framework on the Kinetics-Skeleton and NTU RGB+D 60 datasets, and the best performance is obtained when joint and bone streams are fused (see \tabref{tab4}). After the joint flow and the bone flow are fused, the model can obtain different action information contained in the skeleton data. Compared with the previous single-stream framework, this method can extract more motion information.

\begin{table}[!h]
\caption{The performance comparison of DHGCN with different input data.}\label{tab5}
\begin{tabular*}{\hsize}{@{}@{\extracolsep{\fill}}lcccc@{}}
\toprule
\multirow{2}{*}{Methods} & \multicolumn{2}{c}{Kinetics-Skeleton}& \multicolumn{2}{c}{NTU RGB+D 60}  \\ \cline{2-5}
&Top1&Top5&X-Sub&X-View\\
\hline
DHGCN(joint) &  35.9 & 58.0 &88.6 & 94.8\\
DHGCN(bone) &  35.5 & 58.2 &89.0 & 94.5\\
DHGCN &  \textbf{37.7} & \textbf{60.6} &\textbf{90.7} & \textbf{96.0}\\
\bottomrule
\end{tabular*}
\end{table}

\begin{table}[!h]
\caption{The performance comparison of DHGCN with state-of-the-art methods on the Kinetics-Skeleton dataset.}\label{tab6}
\begin{tabular*}{\hsize}{@{}@{\extracolsep{\fill}}lcc@{}}
\toprule
\multirow{2}{*}{Methods} &  \multicolumn{2}{c}{Kinetics-Skeleton}  \\ \cline{2-3}
&Top1&Top5\\
\hline
TCN \cite{ref_8}  & 20.3 & 40.0\\
\hline
ST-GCN \cite{ref_17} &30.7  & 52.8 \\
ST-GR \cite{ref_24} &33.6  & 56.1 \\
2s-AGCN \cite{ref_18} &36.1  & 58.7 \\
DGNN \cite{ref_22} &36.9  & 59.6 \\
ST-TR \cite{ref_35} &37.4  & 59.8 \\
Advanced CA-GCN \cite{ref_19} &34.1  & 56.6 \\
\hline
DHGCN(Ours) & \textbf{37.7} & \textbf{60.6} \\
\bottomrule
\end{tabular*}
\end{table}

\begin{table}[!h]
\caption{The performance comparison of DHGCN with state-of-the-art methods on the NTU RGB+D 60 dataset.}\label{tab7}
\begin{tabular*}{\hsize}{@{}@{\extracolsep{\fill}}lcc@{}}
\toprule
\multirow{2}{*}{Methods} & \multicolumn{2}{c}{NTU RGB+D 60}  \\ \cline{2-3}
&X-Sub&X-View\\
\hline
Lie Group \cite{ref_2} & 50.1& 82.8 \\
\hline
ST-LSTM \cite{ref_12}  & 69.2 &77.7 \\
ARRN-LSTM \cite{ref_25}  & 80.7 &88.8 \\
Ind-RNN \cite{ref_26}  & 81.8 &88.0 \\
\hline
TCN \cite{ref_8}  & 74.3 &83.1 \\
Clips+CNN+MTLN \cite{ref_9}  & 79.6 &84.8 \\
\hline
ST-GCN \cite{ref_17} &81.5&88.3 \\
Advanced CA-GCN \cite{ref_19} &83.5  & 91.4 \\
ST-GR \cite{ref_24} &86.9&92.3 \\
(P+C)net,Traversal \cite{ref_28} &86.1&93.5 \\
2s-AGCN \cite{ref_18} &88.5&95.1 \\
AGC-LSTM \cite{ref_27} &89.2&95.0 \\
DGNN \cite{ref_22} &89.9&96.1 \\
ST-TR \cite{ref_35} &89.3& 96.1 \\
C-MANs \cite{ref_33} &83.7&93.8 \\
Shift-GCN \cite{ref_39} &\textbf{90.7}&\textbf{96.5}\\
\hline
DHGCN(Ours) & \textbf{90.7}  &96.0 \\
\bottomrule
\end{tabular*}
\end{table}

\begin{table}[!h]
\caption{The performance comparison of DHGCN with state-of-the-art methods on the NTU RGB+D 120 dataset.}\label{tab8}
\begin{tabular*}{\hsize}{@{}@{\extracolsep{\fill}}lcc@{}}
\toprule
\multirow{2}{*}{Methods} & \multicolumn{2}{c}{NTU RGB+D 120}  \\ \cline{2-3}
&X-Sub&X-Set\\
\hline
ST-LSTM \cite{ref_12}  & 55.7 &57.9 \\
\hline
AS-GCN+DH-TCN \cite{ref_44}&78.3&79.8 \\
2s-AGCN \cite{ref_18} &82.5&84.2 \\
ST-TR \cite{ref_35} &82.7& 84.7 \\
Shift-GCN \cite{ref_39} &85.9&87.6\\
\hline
DHGCN(Ours) & \textbf{86.0}  &\textbf{87.9} \\
\bottomrule
\end{tabular*}
\end{table}
\subsection{Comparison with the state-of-the-art}
\noindent We compare the proposed DHGCN and other state-of-the-art methods. \tabref{tab6}, \tabref{tab7} and \tabref{tab8} show the results of the comparison. Our model achieves comparable performance on these three datasets, which proves the superiority of our method.

Since these video from the Kinetics dataset are processed by the OpenPose toolbox, the skeleton data in Kinetics-Skeleton dataset has defects, which adversely affect the performance of the model. We show the Top-1 and Top-5 accuracies in \tabref{tab6} to better reflect the performance of DHGCN. The comparison methods include CNN-based method \cite{ref_8} and GCN-based methods \cite{ref_17,ref_18,ref_19,ref_22,ref_24,ref_35}. ST-GCN is the first method based on GCN. Compared with CNN-based method, it achieves more advanced performance, but the graph topology in this model is fixed. To solve this problem, ST-GR and 2s-AGCN can adaptively learn the high-order information among skeleton joints, these two methods improve the performance of the model. However, they only considered the possible relationship between the two joints and ignored the mutual relationship between multiple joints. Our proposed DHGCN can dynamically construct hypergraph to obtain better performance.

For the NTU RGB+D 60 dataset, the existing methods have achieved advanced performance, therefore, we report the Top-1 accuracy in \tabref{tab7}. The comparison methods include hand-crafted feature methods \cite{ref_2}, RNN-based methods \cite{ref_12,ref_25,ref_26}, CNN-based methods \cite{ref_8,ref_9} and GCN-based methods \cite{ref_17,ref_18,ref_19,ref_22,ref_24,ref_27,ref_28,ref_33,ref_35,ref_39}. 
The X-Sub accuracy of our model exceeds the other fourteen methods and the X-View accuracy is only 0.2\% lower than ST-TR and DGNN. ST-TR introduces self-attention mechanism and proposes spatial and temporal transformer networks. DGNN uses directed acyclic graph to model the skeleton data. These two methods have a slight advantage on the X-View benchmark, but the performance on the X-Sub benchmark is inferior to DHGCN.

NTU RGB+D 120 is a new large-scale dataset. There are relatively little research on this dataset at present. The comparison methods include RNN-based methods \cite{ref_12} and GCN-based methods \cite{ref_44,ref_18,ref_35}. Although our DHGCN performs slightly worse than that of Shift-GCN on the X-View benchmark of the NTU RGB+D 60 dataset, it outperforms Shift-GCN on the NTU RGB+D 120 dataset. Other comparable results can be seen in \tabref{tab8}.

\section{Conclusions and future work}\label{sec:conclusions}

\noindent In this paper, we propose a dynamic hypergraph convolutional network (DHGCN) for skeleton-based action recognition. DHGCN uses a combination of static hypergraph and dynamic hypergraph to model the skeleton data, which can extract more information than traditional skeleton graph. The dynamic hypergraph is composed of two important mechanisms: dynamic joint weighting and dynamic topology. Dynamic joint weighting assigns different weights to each node according to its moving distance, which can better help to aggregate the node features. Dynamic topology improves the flexibility of the model by adjusting the topology of the skeleton hypergraph. Moreover, the introduction of joint-bone fusion framework further improves the performance. The proposed model is evaluated on the Kinetics-Skeleton, NTU RGB+D 60 and NTU RGB+D 120 datasets, and the performance of our model achieves competitive performance on these three datasets. 

Although DHGCN achieves good performance, we also find some problems that need to be resolved. Our model contains a ten-layer convolutional network, and the network has complex calculations in the process of obtaining dynamic hypergraph. Our method mainly focuses on the construction of human skeleton topology, but it ignores the influence of scene factors. For example, when someone makes a phone call, not only joint motion information needs to be considered, but also background factors such as mobile phone needs to be considered. In the future, we will continue to optimize the model to reduce network depth and computational complexity. Furthermore, we consider to fuse the skeleton data with the RGB data so that the model can use the joint motion information and the scene factors from the video at the same time.


\CvmAck{The authors would like to thank all the anonymous reviewers. This work was supported in part by the National Natural Science Foundation of China under Grant No.61772474, No.62036010; in part by the Program for Science and Technology Innovation Talents in Universities of Henan Province under Grant 20HASTIT021.}

\bibliographystyle{CVM}

{\normalsize  \bibliography{ref}}

~\\
~\\
~\\
~\\
~\\
~\\
~\\
~\\

\Author{cvm/author1.jpg}{Jinfeng Wei}
{received his B.E. degree from the School of Information Engineering at Zhengzhou University in 2019. He is currently working toward his M.S. degree at Henan Institute of Advanced Technology, Zhengzhou University. His research interests include computer vision and computer graphics.}

\Author{cvm/author2.jpg}{Yunxin Wang}
{received his B.E.  degree from the School of college of science at Northeast Electric Power University in 2018. He is currently working toward his M.E. degree at Henan Institute of Advanced Technology, Zhengzhou University. His research interests include graph neural network and deep learning.  }

\Author{cvm/author3.jpg}{Mengli Guo}
{received her B.E. degree from the School of Information Engineering at Zhengzhou University in 2019. She is currently working toward her M.S. degree in School of Computer and Artificial Intelligence, Zhengzhou University. Her research interests include computer vision and computer graphics. }

\Author{cvm/author4.jpg}{Pei Lv}
{received the Ph.D. degree from the State Key Laboratory of CAD\&CG, Zhejiang University Hangzhou, China, in 2013. He is an Associate Professor with the School of Computer and Artificial Intelligence, Zhengzhou University, Zhengzhou, China. His research interests include computer vision and computer graphics. He has authored more than 30 journal and conference papers in the above areas, including the IEEE TRANSACTIONS ON IMAGE PROCESSING , the IEEE TRANSACTIONS ON CIRCUITS AND SYSTEMS FOR VIDEO TECHNOLOGY, CVPR, ACM MM, and IJCAI.}


\Author{cvm/author5.jpg}{Xiaoshan Yang}
{received the master's degree in computer science from Beijing Institute of Technology, Beijing, China, in 2012, and the Ph.D. degree in pattern recognition and intelligent systems from the Institute of Automation, Chinese Academy of Sciences, Beijing, China, in 2016. Currently, he is an Associate Professor with the Institute of Automation, Chinese Academy of Sciences, Beijing, China. His research interests include multimedia analysis and computer vision.}

\Author{cvm/author6.jpg}{Mingliang Xu}
{received the Ph.D. degree in computer science and technology from the State Key Laboratory of CAD\&CG, Zhejiang University, Hangzhou, China, in 2012. He is a Full Professor with the School of Computer and Artificial Intelligence, Zhengzhou University, Zhengzhou, China, where he is currently the Director of the Center for Interdisciplinary Information Science Research and the Vice General Secretary of ACM SIGAI China. His research interests include computer graphics, multimedia, and artificial intelligence. He has authored more than 60 journal and conference papers in the above areas, including the ACM Transactions on Graphics, the ACM Transactions on Intelligent Systems and Technology, 
the IEEE TRANSACTIONS ON PATTERN ANALYSIS AND MACHINE INTELLIGENCE, the IEEE TRANSACTIONS ON IMAGE PROCESSING, 
the IEEE TRANSACTIONS ON CYBERNETICS, 
the IEEE TRANSACTIONS ON CIRCUITS AND SYSTEMS FOR VIDEO TECHNOLOGY, ACM SIGGRAPH (Asia), ACM MM, and ICCV.}

\end{document}